\documentclass[10pt,twocolumn,letterpaper]{article}

\usepackage[pagenumbers]{cvpr} 

\usepackage{graphicx}
\usepackage{amsmath}
\usepackage{amssymb}
\usepackage{booktabs}
\usepackage{xcolor}

\usepackage{float}
\restylefloat{table}
\usepackage{adjustbox}

\usepackage{bm}
\usepackage{enumitem}
\usepackage{comment}

\usepackage[pagebackref,breaklinks,colorlinks]{hyperref}

\usepackage[capitalize]{cleveref}
\crefname{section}{Sec.}{Secs.}
\Crefname{section}{Section}{Sections}
\Crefname{table}{Table}{Tables}
\crefname{table}{Tab.}{Tabs.}

\def\cvprPaperID{1749} 
\def\confName{CVPR}
\def\confYear{2023}

\makeatletter
\renewcommand{\paragraph}{%
  \@startsection{paragraph}{4}%
  {\z@}{0.8ex \@plus 0.2ex \@minus .2ex}{-1em}%
  {\normalfont\normalsize\bfseries}%
}
\makeatother

\begin{document}

\setlength{\abovedisplayskip}{5pt}
\setlength{\belowdisplayskip}{5pt}
\setlength\abovedisplayshortskip{0pt}

\title{Efficient View Synthesis and 3D-based Multi-Frame Denoising\\ with Multiplane Feature Representations}

\author{Thomas Tanay \quad\quad Aleš Leonardis \quad\quad Matteo Maggioni\\
Huawei Noah's Ark Lab\\
{\tt\small \{thomas.tanay,ales.leonardis,matteo.maggioni\}@huawei.com}
}
\maketitle

\begin{abstract}
While current multi-frame restoration methods combine information from multiple input images using 2D alignment techniques, recent advances in novel view synthesis are paving the way for a new paradigm relying on volumetric scene representations. In this work, we introduce the first 3D-based multi-frame denoising method that significantly outperforms its 2D-based counterparts with lower computational requirements. Our method extends the multiplane image (MPI) framework for novel view synthesis by introducing a learnable encoder-renderer pair manipulating multiplane representations in feature space. The encoder fuses information across views and operates in a depth-wise manner while the renderer fuses information across depths and operates in a view-wise manner. The two modules are trained end-to-end and learn to separate depths in an unsupervised way, giving rise to Multiplane Feature (MPF) representations. Experiments on the Spaces and Real Forward-Facing datasets as well as on raw burst data validate our approach for view synthesis, multi-frame denoising, and view synthesis under noisy conditions.
\end{abstract}

\section{Introduction}
\label{sec:intro}

Multi-frame denoising is a classical problem of computer vision where a noise process affecting a set of images must be inverted. The main challenge is to extract consistent information across images effectively and the current state of the art relies on optical flow-based 2D alignment~\cite{tian2020tdan, chan2022basicvsr++, bhat2021deep}. Novel view synthesis, on the other hand, is a classical problem of computer graphics where a scene is viewed from one or more camera positions and the task is to predict novel views from target camera positions. This problem requires to reason about the 3D structure of the scene and is typically solved using some form of volumetric representation~\cite{penner2017soft,zhou2018stereo,mildenhall2020nerf}. Although the two problems are traditionally considered distinct, some novel view synthesis approaches have recently been observed to handle noisy inputs well, and to have a denoising effect in synthesized views by discarding inconsistent information across input views~\cite{kalantari2016learning,mildenhall2022nerf}. This observation opens the door to 3D-based multi-frame denoising, by recasting the problem as a special case of novel view synthesis where the input views are noisy and the target views are the clean input views~\cite{mildenhall2022nerf,pearl2022nan}.  

\begin{figure}[t]
  \centering
    \includegraphics[width=\linewidth]{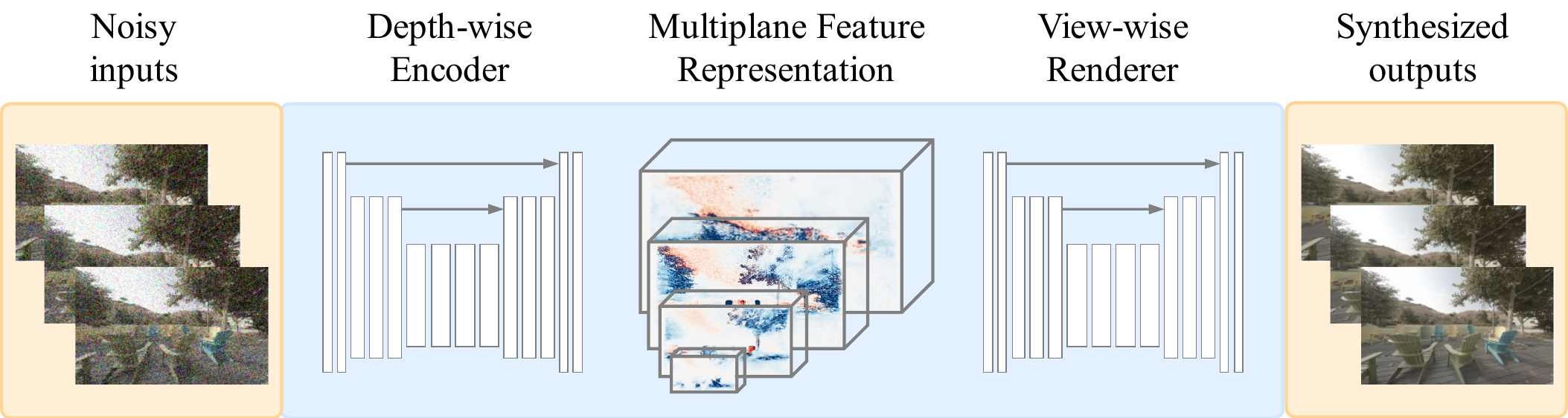}
    \makebox[0.24\linewidth]{\scriptsize Noisy inputs}\hfill
    \makebox[0.24\linewidth]{\scriptsize IBRNet~\cite{wang2021ibrnet}}\hfill
    \makebox[0.24\linewidth]{\scriptsize NAN~\cite{pearl2022nan}}\hfill
    \makebox[0.24\linewidth]{\scriptsize MPFER (ours)}
    \includegraphics[width=0.24\linewidth]{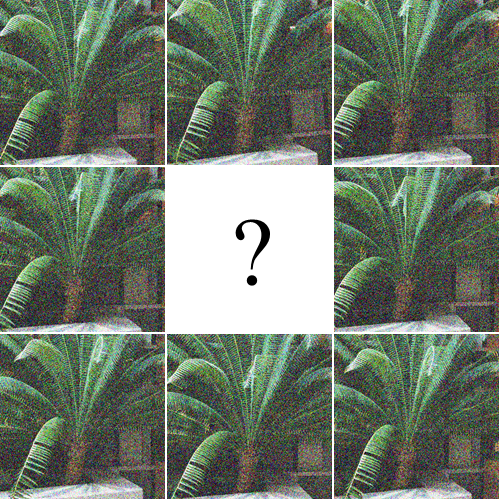}\hfill
    \includegraphics[width=0.24\linewidth]{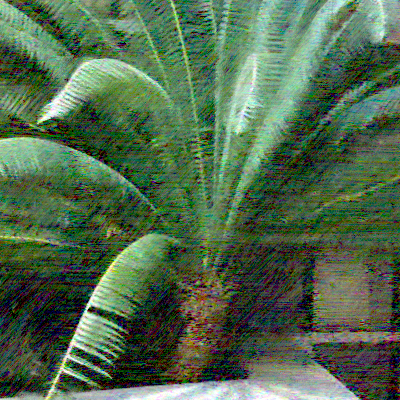}\hfill
    \includegraphics[width=0.24\linewidth]{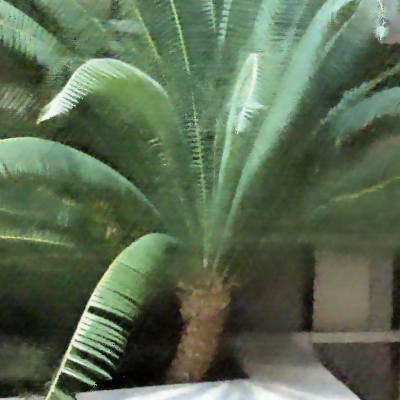}\hfill \includegraphics[width=0.24\linewidth]{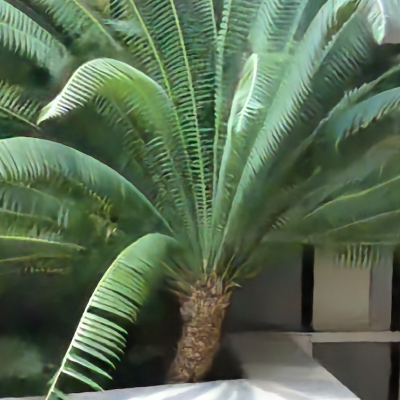}\vspace{-0.05cm}
  \caption{\emph{Top:} Our Multiplane Features Encoder-Renderer (MPFER) reimagines the MPI pipeline by moving the multiplane representation to feature space. \emph{Bottom:} MPFER significantly outperforms existing methods in multiple challenging scenarios, including here, novel view synthesis from 8 highly degraded inputs.}
  \label{fig:teaser}
\end{figure}

Recently, novel view synthesis has been approached as an encoding-rendering process where a scene representation is first \emph{encoded} from a set of input images and an arbitrary number of novel views are then \emph{rendered} from this scene representation. In the Neural Radiance Field (NeRF) framework for instance, the scene representation is a radiance field function encoded by training a neural network on the input views. Novel views are then rendered by querying and integrating this radiance field function over light rays originating from a target camera position~\cite{mildenhall2020nerf,barron2021mip,martin2021nerf}. In the Multiplane Image (MPI) framework on the other hand, the scene representation is a stack of semi-transparent colored layers arranged at various depths, encoded by feeding the input views to a neural network trained on a large number of scenes. Novel views are then rendered by warping and overcompositing the semi-transparent layers~\cite{szeliski1998stereo,zhou2018stereo,flynn2019deepview}. 

In the present work, we adopt the MPI framework because it is much lighter than the NeRF framework computationally. The encoding stage only requires one inference pass on a network that generalizes to new scenes instead of training one neural network per-scene, and the rendering stage is essentially free instead of requiring a large number of inference passes. However, the standard MPI pipeline struggles to predict multiplane representations that are self-consistent across depths from multiple viewpoints. This problem can lead to depth-discretization artifacts in synthesized views~\cite{srinivasan2019pushing} and has previously been addressed at the encoding stage using computationally expensive mechanisms and a large number of depth planes~\cite{srinivasan2019pushing,mildenhall2019local,flynn2019deepview,han2022single}. Here, we propose to enforce cross-depth consistency at the rendering stage by replacing the fixed overcompositing operator with a learnable renderer. This change of approach has three important implications. First, the encoder module can now process depths independently from each other and focus on fusing information across views. This significantly reduces the computational load of the encoding stage. Second, the scene representation changes from a static MPI to Multiplane Features (MPF) rendered dynamically. This significantly increases the expressive power of the scene encoding. Finally, the framework's overall performance is greatly improved, making it suitable for novel scenarios including multi-frame denoising where it outperforms standard 2D-based approaches at a fraction of their computational cost. Our main contributions are as follow:

\begin{itemize}[leftmargin=*,noitemsep,topsep=0pt]
    \item We solve the cross-depth consistency problem for multiplane representations at the rendering stage, by introducing a learnable renderer.
    \item We introduce the Multiplane Feature (MPF) representation, a generalization of the multiplane image with higher representational power.
    \item We re-purpose the multiplane image framework originally developed for novel view synthesis to perform 3D-based multi-frame denoising.
    \item We validate the approach with experiments on 3 tasks and 3 datasets and significantly outperform existing 2D-based and 3D-based methods for multi-frame denoising. 
\end{itemize}

\section{Related work}
\label{sec:related work}

\paragraph{Multi-frame denoising} 
Multi-frame restoration methods are frequently divided into two categories, depending on the type of image alignment employed. \emph{Explicit alignment} refers to the direct warping of images using optical flows predicted by a motion compensation module~\cite{caballero2017real,tao2017detail,tassano2019dvdnet,xue2019video}. In contrast, \emph{implicit alignment} refers to local, data-driven deformations implemented using dynamic upsampling filters~\cite{jia2016dynamic,jo2018deep}, deformable convolutions~\cite{tian2020tdan,wang2019edvr}, kernel prediction networks~\cite{mildenhall2018burst} or their extension, basis prediction networks~\cite{xia2020basis}. Explicit alignment is better at dealing with large motion while implicit alignment is better at dealing with residual motion, and state-of-the-art performance can be achieved by combining both in the form of flow-guided deformable convolutions~\cite{chan2021understanding,chan2022basicvsr++}. Another distinction between multi-frame restoration methods is the type of processing used. A common approach is to concatenate the input frames together along the channel dimension \cite{caballero2017real,tao2017detail,tassano2019dvdnet,xue2019video,wang2019edvr} but recurrent processing is more efficient~\cite{godard2018deep,sajjadi2018frame,fuoli2019efficient,tanay2022diagnosing}, especially when implemented in a bidirectional way~\cite{huang2015bidirectional,chan2021basicvsr,chan2022basicvsr++}. BasicVSR++ achieves state-of-the-art performance by combining flow-guided deformable alignment with bidirectional recurrent processing iterated multiple times~\cite{chan2022basicvsr++}. In a different spirit, the recent DeepRep method~\cite{bhat2021deep} introduces a deep reparameterization in feature space of the maximum a posteriori formulation of multi-frame restoration. Similarly to the previous methods however, it still uses a form of explicit 2D alignment, and lacks any ability to reason about the 3D structure of the scene. 

\paragraph{View synthesis} 
The idea to decompose a scene into a set of semi-transparent planes can be traced back to the use of mattes and blue screens in special effects film-making~\cite{vlahos1993traveling,smith1996blue}. This scene representation was first applied to view interpolation in~\cite{szeliski1998stereo}, and recently gained popularity under the name of Multiplane Image (MPI)~\cite{zhou2018stereo}. It is particularly powerful to generate novel views from a small set of forward facing views~\cite{zhou2018stereo,mildenhall2019local,flynn2019deepview}, and can even be used to generate novels views from a single image~\cite{tucker2020single,li2021mine,han2022single}. The rendering of view dependent-effects and non-Lambertian surfaces is challenging due to the use of a single set of RGB$\alpha$ images, and can be improved by predicting multiple MPIs combined as a weighted average of the distance from the input views~\cite{mildenhall2019local}, or as a set of basis components~\cite{wizadwongsa2021nex}. The simplicity of this representation is appealing, but it can still be computationally heavy when the number of depth planes grows~\cite{flynn2019deepview,wizadwongsa2021nex}, and the rendered views can suffer from depth discretization artifacts~\cite{srinivasan2019pushing}. A number of alternative layered scene representations exist, including the Layered Depth Image (LDI) consisting in one RGB image with an extra depth channel~\cite{shade1998layered}, and variants of MPIs and LDIs~\cite{lin2020deep,hu2021worldsheet,khakhulin2022stereo,solovev2023self}. So far however, all these methods use a fixed overcompositing operator at the rendering stage. The idea to perform view synthesis by applying 3D operations in the feature space of an encoder-decoder architecture was explored on simple geometries in~\cite{worrall2017interpretable}, and used with success on a point-cloud scene representation for view synthesis from a single image in~\cite{wiles2020synsin}. Recently, Neural Radiance Fields (NeRFs) have become highly popular for their ability to produce high quality renderings of complex scenes from arbitrary viewpoints~\cite{mildenhall2020nerf,barron2021mip,martin2021nerf,niemeyer2022regnerf,park2021nerfies}. However, they tend to be very heavy computationally, require a large number of input views, and lack the ability to generalize to novel scenes. IBRNet~\cite{wang2021ibrnet} improves generalizability by learning a generic view interpolation function, but the approach remains computationally heavy. The application of view synthesis approaches to multi-frame restoration has been limited so far, and exclusively based on NeRF. RawNeRF~\cite{mildenhall2022nerf} explores novel view synthesis in low-light conditions, and reports strong denoising effects in the synthesized views. Deblur-NeRF~\cite{ma2022deblur} augments the ability of NeRFs to deal with blur by including Deformable Sparse Kernels. Noise-aware-NeRFs~\cite{pearl2022nan} improves the IBRNet architecture to explicitly deal with noise. However, these different restoration approaches still suffer from the limitations affecting their underlying NeRF representations.

\section{Method}

We start by describing the standard MPI processing pipeline, before discussing the cross-depth consistency problem. We then introduce our MPF Encoder-Renderer and its adaptation to multi-frame denoising.

\subsection{Standard MPI processing pipeline}

\begin{figure}[t]
  \centering
  \begin{subfigure}{\linewidth}
    \includegraphics[width=\linewidth]{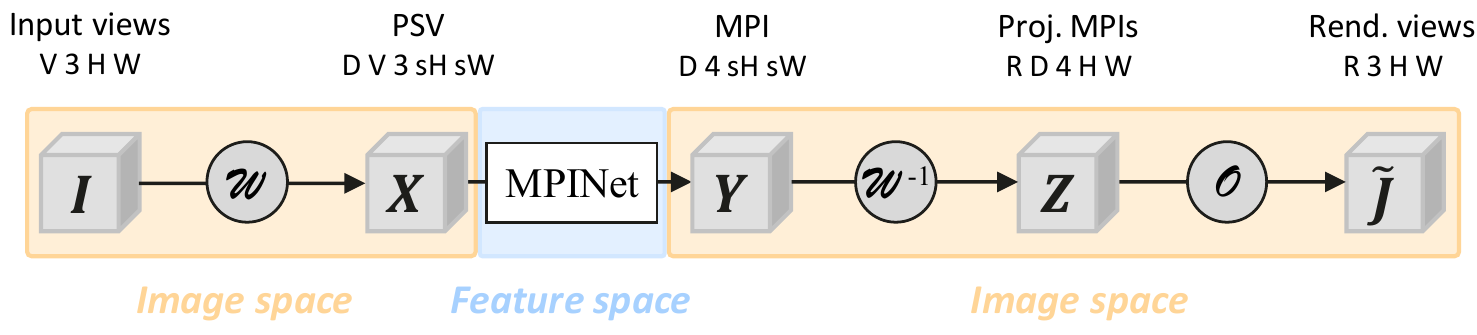}
    \caption{Standard MPI processing pipeline.}
    \label{fig:comparison-a}
  \end{subfigure}
  \par\bigskip
  \begin{subfigure}{\linewidth}
    \includegraphics[width=\linewidth]{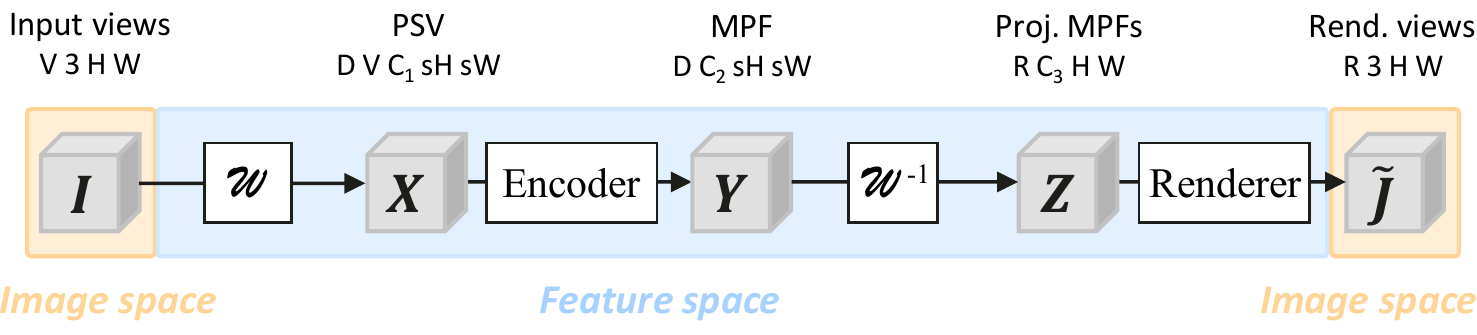}
    \caption{Our MPF Encoder-Renderer.}
    \label{fig:comparison-b}
  \end{subfigure}
   \caption{In the standard MPI processing pipeline, all the learning and most of the processing happens in the MPINet module. We propose to move the multiplane representation to feature space, by giving some processing power to the warping operators, and replacing the overcompositing operator with a learnable renderer.}
   \label{fig:comparison}
\end{figure}

The standard MPI processing pipeline turns a set of input views into an arbitrary set of rendered novel views by applying 4 main transformations: forward-warping, MPI prediction, backward-warping, overcompositing (See \Cref{fig:comparison-a}). We describe this pipeline in more details below.

\paragraph{Input views} The inputs of the pipeline are a set of $V$ views of a scene, consisting of images and camera parameters. The images are of height $H$ and width $W$, with red-green-blue color channels, and can be stacked into a 4D tensor $\bm{I} = \{\{\{\{\bm{I}_{vchw}\}_{w=1}^W\}_{h=1}^H\}_{c=1}^3\}_{v=1}^V$. To simplify notations, we omit the dimensions $c,h,w$ and refer to an individual image as $\bm{I}_v$. The camera parameters consist of an intrinsic tensor $\bm{K}$ of size $V\!\times\!3\!\times\!3$ containing information about the focal lengths and principal point of the cameras, and an extrinsic tensor containing information about the camera orientations in the world coordinate system, that can be split into a rotation tensor~$\bm{R}$ of size $V\!\times\!3\!\times\!3$ and a translation tensor~$\bm{t}$ of size $V\!\times\!3\!\times\!1$. A reference view~$i$ is defined a priori, and the positions of all the cameras are assumed to be expressed relatively to it. The intrinsic matrix $\bm{K}_i$ of the reference camera is defined such that the corresponding field of view covers the region of space visible to all the input views. Finally, a set of $D$ depth planes is distributed orthogonally to the reference viewing direction such that their normal is $\bm{n} = (0, 0, 1)^\top$, and their distances $\{a_d\}_{d=1}^D$ from the reference camera center are sampled uniformly in disparity. The camera parameters and the depth planes are used to define a set of $D\!\times\!V$ homography projections, represented by a tensor $\bm{H}$ of size $D\!\times\!V\!\times\!3\!\times\!3$. Each homography is between one of the input views and the reference view, and is induced by one of the depth planes, such that its matrix is expressed as~\cite{hartley2003multiple}:
\begin{equation}
\bm{H}_{dv} = \bm{K}_v \left( \bm{R}_v - \frac{\bm{t}_v \, \bm{n}^\top}{a_d} \right) \bm{K}_i^{-1}
\label{eq:homography}
\end{equation}

\paragraph{Plane Sweep Volumes}
The first transformation in the MPI pipeline is the computation of Plane Sweep Volumes (PSVs), obtained by forward-warping each input image $D$ times, according to the homography $\bm{H}_{dv}$. The sampling rate of the warping operator $\mathcal{W}$ is a hyperparameter that can be controlled using an up-scaling factor $s$. Each transformation can thus be written as $\bm{X}_{dv} = \mathcal{W}( \bm{I}_{v} ,\, \bm{H}_{dv} ,\, s)$ and the PSV tensor $\bm{X}$ is of size $D\!\times\!V\!\times\!3\!\times\!sH\!\times\!sW$.

\paragraph{Multiplane Image}
The main processing block of the pipeline is a neural network $\text{MPINet}$, turning the set of PSVs into a multiplane image representation of the scene $\bm{Y} = \text{MPINet}( \bm{X} )$ where $\bm{Y}$ is a set of $D$ semi-transparent RGB images of size $D\!\times\!4\!\times\!sH\!\times\!sW$, constrained to the [0,1] range by using a sigmoid activation function.

\paragraph{Projected MPIs}
The MPI is then backward-warped to a set of $R$ novel views, defined by an homography tensor $\bm{G}$ of size $R\!\times\!D\!\times\!3\!\times\!3$ following \cref{eq:homography} with the depth and view dimensions transposed. The backward-warping operation is defined as $\bm{Z}_{rd} = \mathcal{W}\left( \bm{Y}_{d} ,\, \bm{G}_{rd}^{-1} ,\, 1/s \right)$ obtaining a tensor of projected MPIs $\bm{Z}$ of size $R\!\times\!D\!\times\!4\!\times\!H\!\times\!W$.

\paragraph{Rendered views}
The projected MPIs are finally collapsed into single RGB images by applying the overcompositing operator $\mathcal{O}$. This operator splits each RGB$\alpha$ image $\bm{Z}_{rd}$ into a colour component $\bm{C}_{rd}$ and an alpha component $\bm{A}_{rd}$ and computes $\bm{\tilde{J}}_r = \sum_{d=1}^{D}\left( \bm{C}_{rd} \bm{A}_{rd} \prod_{k=d+1}^{D} \left( 1 - \bm{A}_{rk} \right) \right)$ obtaining the rendered views $\bm{\tilde{J}}$ of size $R\!\times\!3\!\times\!H\!\times\!W$.

\paragraph{Training}
The pipeline is typically trained end-to-end in a supervised way by minimizing a loss $\mathcal{L}(\bm{\tilde{J}}, \bm{J})$ between the rendered views $\bm{\tilde{J}}$ and the corresponding ground-truth images $\bm{J}$. In practice, $\mathcal{L}$ is often an $\mathcal{L}_1$ loss applied to low level features of a VGG network~\cite{simonyan2014very}.

\subsection{Cross-depth consistency}
\label{sec:Cross-depth consistency}

The main and only learnable module of the standard MPI pipeline is the prediction network MPINet, transforming a 5D PSV tensor into a 4D MPI scene representation. Its task is challenging because multiplane images are hyper-constrained representations: their semi-transparent layers interact with each other in non-trivial and view-dependent ways, and missing or redundant information across layers can result in depth discretization artifacts after applying the overcompositing operator~\cite{srinivasan2019pushing}. To perform well, MPINet requires a mechanism enforcing \emph{cross-depth consistency}; and several approaches have been considered before.

In the original case of stereo-magnification~\cite{zhou2018stereo}, there is one reference input and one secondary input which is turned into a single PSV. Both are concatenated along the channel dimension and fed to an MPINet module predicting the full MPI in one shot as a 3D tensor of size $(D\!\times\!4)\!\times\!sH\!\times\!sW$, such that cross-depth consistency is enforced within the convolutional layers of MPINet. A similar solution can be used in the case of single-view view synthesis~\cite{tucker2020single}, where there is a single reference input image and no PSV. In the general case with $V$ inputs, however, the PSV tensor $\bm{X}$ becomes very large and there are two main ways to process it.
\begin{description}[noitemsep,topsep=1pt,leftmargin=20pt]
    \item[Option 1] The first solution is to generalize the approach of~\cite{zhou2018stereo} and concatenate $\bm{X}$ across views and depths before feeding it to a network predicting the full MPI in one shot: $\bm{Y} = \text{MPINet}\!\left( \left\{\{\bm{X}_{dv}\}_{v=1}^V\right\}_{d=1}^D \right)$.
    \item[Option 2] The second solution is to concatenate $\bm{X}$ across views, and process each depth separately---effectively running the MPINet block $D$ times in parallel:
    \mbox{$\bm{Y} = \left\{\text{MPINet}\!\left( \{\bm{X}_{dv}\}_{v=1}^V \right)\right\}_{d=1}^D$}.
\end{description}
Option~1 tends to work poorly in practice~\cite{flynn2019deepview}, as it requires to either use very large convolutional layers with intractable memory requirements, or discard most of the information contained in the input PSVs after the first convolutional layer. Option~2 is appealing as it fuses information across views more effectively, but the resulting MPI typically suffers from a lack of cross-depth consistency as each depth is processed separately. Most previous works adopt Option~2 as a starting point, and augment it with various mechanisms allowing some information to flow across depths. For instance, some methods implement MPINet with 3D convolutions~\cite{srinivasan2019pushing,mildenhall2019local}, such that each depth is treated semi-independently within a local depth neighborhood dependent on the size of the kernel, which is typically 3. By design however, this solution cannot handle interactions between distant depth planes. DeepView~\cite{flynn2019deepview} proposes to solve the recursive cross-depth constraint by iteratively refining the prediction of the MPINet block, effectively performing a form of learned gradient descent. However, this solution requires to run the MPI network multiple times, which is computationally heavy. Finally, the method of~\cite{han2022single} uses a feature masking strategy to deal with inter-plane interactions explicitly. This solution is both complex (multiple networks are required to predict the masks) and rigid (the masking operations are fixed and still work on a per-depth basis). 

\begin{figure*}[t]
  \centering
   \includegraphics[width=0.88\linewidth]{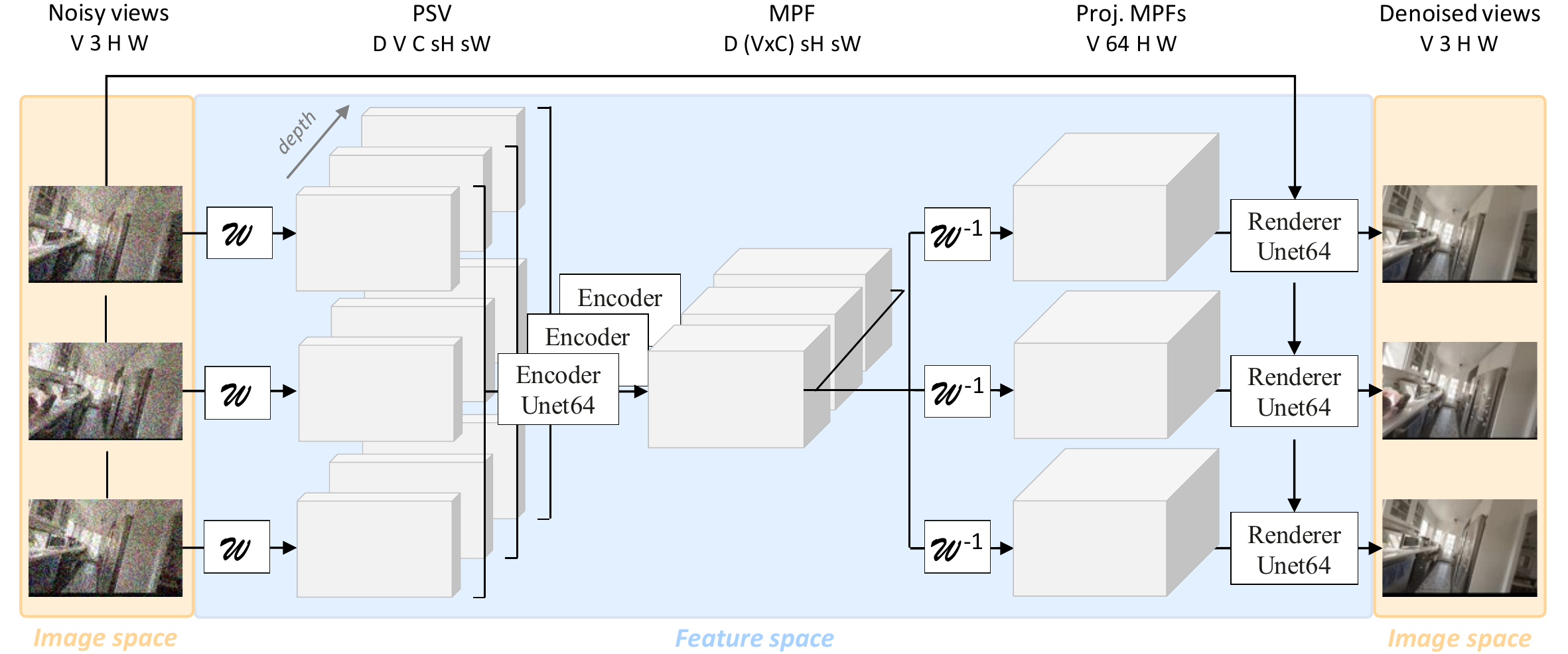}

   \caption{Our Multiplane Features Encoder-Renderer (MPFER). Input views are forward-warped into plane sweep volumes (PSVs) which are processed depthwise by the Encoder Unet64. The resulting multiplane feature representation (MPF) can then be back-projected to an arbitrary number of novel views, or to the same views as the inputs---allowing the integration of a skip connection (illustrated here). The Renderer Unet64 processes the projected MPFs on a per-view basis, producing the final synthesised or denoised outputs.}
   \label{fig:implementation}
\end{figure*}

\subsection{Our MPF Encoder-Renderer}

Here, we propose to solve the cross-depth consistency problem in a novel way, by addressing it at the rendering stage. Specifically, we replace the fixed overcompositing operator with a learnable renderer enforcing consistency directly at the output level on a per-view basis. This design change greatly simplifies the task of the multiplane representation encoder, which can now focus on fusing information across views in a depth-independent way. It also promotes the multiplane representation to feature space, by relaxing existing constraints on the number of channels and scaling in the [0,1] range. The four transformations of the pipeline are modified as follows (See \Cref{fig:comparison-b}).

\paragraph{Plane Sweep Volumes} 
To decrease the amount of information loss through image warping and promote the PSVs to feature space, we now apply a convolution to the images before warping them: $\bm{X}_{dv} = \mathcal{W}( \text{Conv}(\bm{I}_{v}) ,\, \bm{H}_{dv} ,\, s)$. The PSV tensor $\bm{X}$ is now of size $D\!\times\!V\!\times\!C_1\!\times\!sH\!\times\!sW$ where the number of channels $C_1$ is a hyperparameter.

\paragraph{Multiplane Features}
We then replace the MPINet module with an encoder applied to each depth independently: $\bm{Y} = \left\{\text{Encoder}\left( \{\bm{X}_{dv}\}_{v=1}^V \right)\right\}_{d=1}^D$. The multiplane representation $\bm{Y}$ is now in feature space: its range is not constrained by a sigmoid function anymore and its size is $D\!\times\!C_2\!\times\!sH\!\times\!sW$ where the number of channels $C_2$ is a second hyperparameter.

\paragraph{Projected MPFs}
The MPF is still backward-warped to a set of $R$ novel views defined by an homography tensor $\bm{G}$, but this is now followed by a convolution collapsing the depth and channel dimensions into a single dimension: $\bm{Z}_{r} = \text{Conv}\!\left(\left\{\mathcal{W}\left( \bm{Y}_{d} ,\, \bm{G}_{dr}^{-1} ,\, 1/s \right)\right\}_{d=1}^D\right)$. The tensor $\bm{Z}$ is of size $R\!\times\!C_3\!\times\!H\!\times\!W$ where $C_3$ is a third hyperparameter.

\paragraph{Rendered views}
Finally, the projected MPFs are turned into the final rendered views by a simple CNN renderer operating on each view separately: $\bm{\tilde{J}} = \left\{\text{Renderer}( \bm{Z}_{r} )\right\}_{r=1}^R$. The rendered views $\bm{\tilde{J}}$ are of size $R\!\times\!3\!\times\!H\!\times\!W$.

\paragraph{Training}
The pipeline is still trained end-to-end by minimizing a loss $\mathcal{L}(\bm{\tilde{J}}, \bm{J})$ between the rendered views $\bm{\tilde{J}}$ and the corresponding ground-truth images $\bm{J}$, but the renderer and the warping operators are now learnable. When the input views $\bm{I}$ are noisy versions of ground-truth images $\bm{I}^\ast$, the pipeline can be turned into a multi-frame denoising method by using the input homographies $\bm{H}$ during backward-warping, obtaining denoised outputs $\bm{\tilde{I}}$, and by minimizing the loss $\mathcal{L}(\bm{\tilde{I}}, \bm{I}^\ast)$. There is then a one-to-one mapping between the input views and the rendered views, and it is possible to integrate a skip connection feeding the noisy inputs directly to the renderer to guide its final predictions. In all our experiments, we use Unets~\cite{ronneberger2015u} with a base of 64 channels to implement both the encoder and the renderer. We set $C_1 = C$, $C_2 = V\times C$ and $C_3 = 64$ such that there is a single hyperparameter $C$ to vary. Our Multiplane Features Encoder-Render (MPFER) is illustrated in \Cref{fig:implementation} for 3 inputs views and 3 depth planes. A MindSpore~\cite{mindspore} implementation of our method is available\footnote{\url{https://github.com/mindspore-lab/mindediting}}.

\begin{figure*}
  \centering
    \makebox[0.163\linewidth]{\footnotesize Noisy input}\hfill
    \makebox[0.163\linewidth]{\footnotesize BPN~\cite{xia2020basis}}\hfill
    \makebox[0.163\linewidth]{\footnotesize BasicVSR++~\cite{chan2022basicvsr++}}\hfill
    \makebox[0.163\linewidth]{\footnotesize DeepRep~\cite{bhat2021deep}}\hfill
    \makebox[0.163\linewidth]{\footnotesize MPFER-64 (ours)}\hfill
    \makebox[0.163\linewidth]{\footnotesize Ground truth}
    \includegraphics[width=0.163\linewidth]{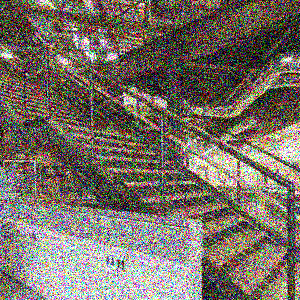}\hfill \includegraphics[width=0.163\linewidth]{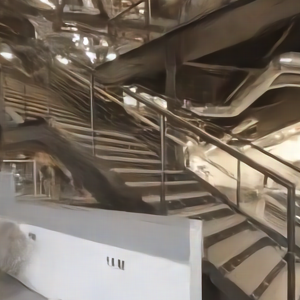}\hfill \includegraphics[width=0.163\linewidth]{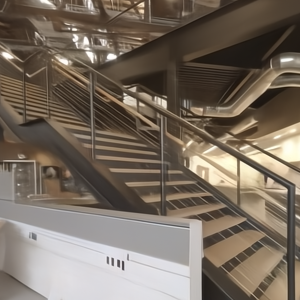}\hfill \includegraphics[width=0.163\linewidth]{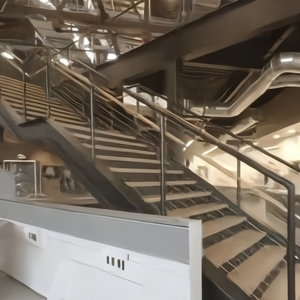}\hfill \includegraphics[width=0.163\linewidth]{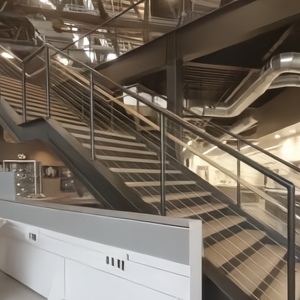}\hfill \includegraphics[width=0.163\linewidth]{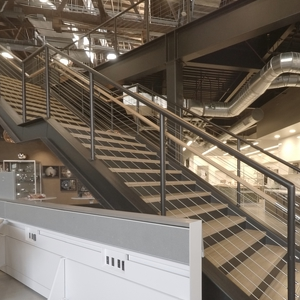}\vspace{0.05cm}
    \includegraphics[width=0.163\linewidth]{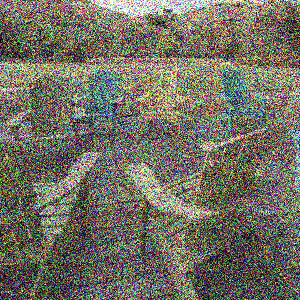}\hfill \includegraphics[width=0.163\linewidth]{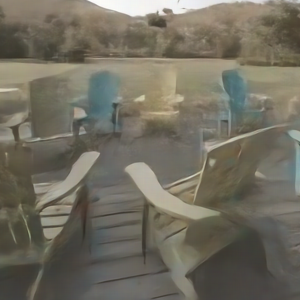}\hfill \includegraphics[width=0.163\linewidth]{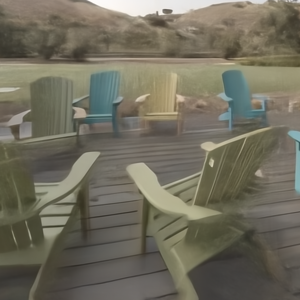}\hfill \includegraphics[width=0.163\linewidth]{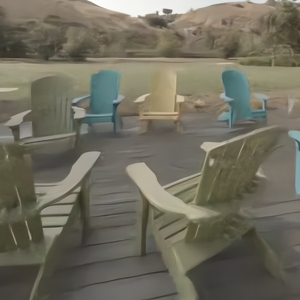}\hfill \includegraphics[width=0.163\linewidth]{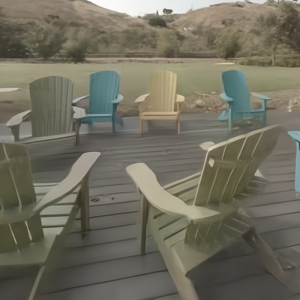}\hfill \includegraphics[width=0.163\linewidth]{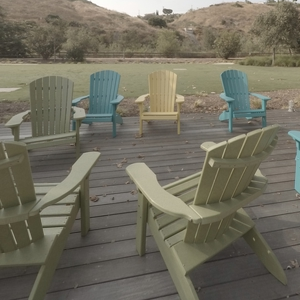}
    \makebox[0.163\linewidth]{\footnotesize Noisy input}\hfill 
    \makebox[0.163\linewidth]{\footnotesize IBRNet-N~\cite{pearl2022nan}}\hfill
    \makebox[0.163\linewidth]{\footnotesize NAN~\cite{pearl2022nan}}\hfill
    \makebox[0.163\linewidth]{\footnotesize MPFER-N (ours)}\hfill
    \makebox[0.163\linewidth]{\footnotesize MPFER-C (ours)}\hfill
    \makebox[0.163\linewidth]{\footnotesize Ground truth}
    \includegraphics[width=0.163\linewidth]{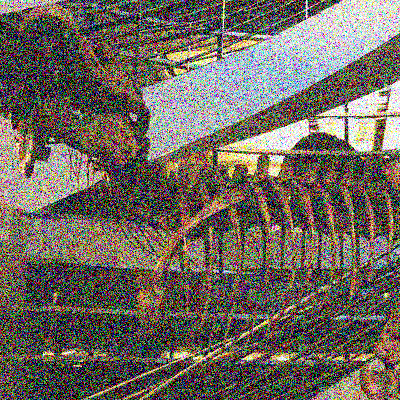}\hfill \includegraphics[width=0.163\linewidth]{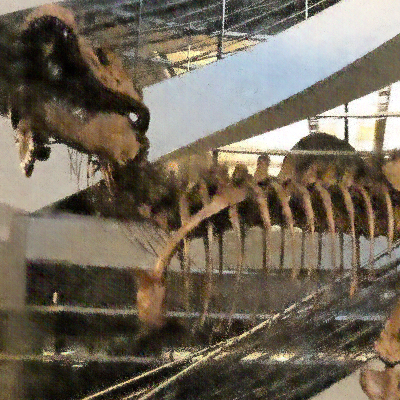}\hfill \includegraphics[width=0.163\linewidth]{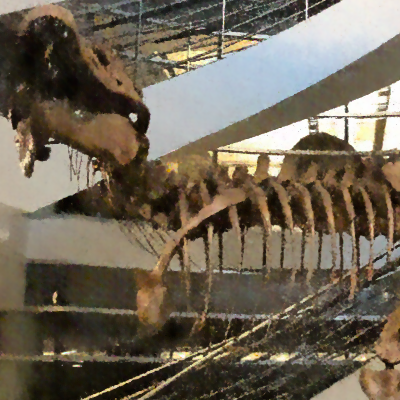}\hfill \includegraphics[width=0.163\linewidth]{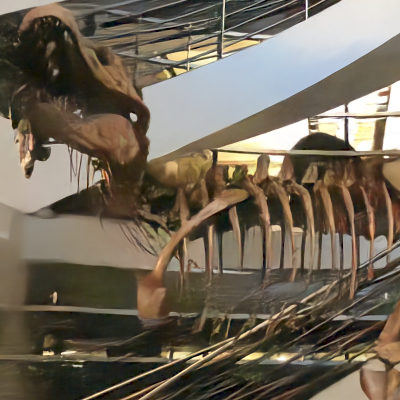}\hfill \includegraphics[width=0.163\linewidth]{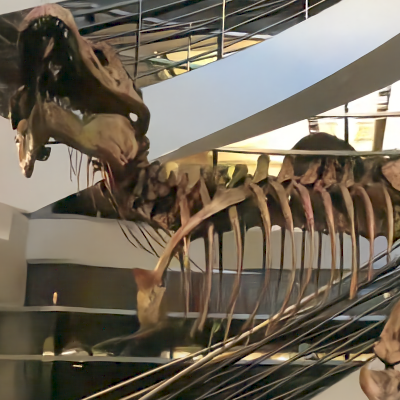}\hfill \includegraphics[width=0.163\linewidth]{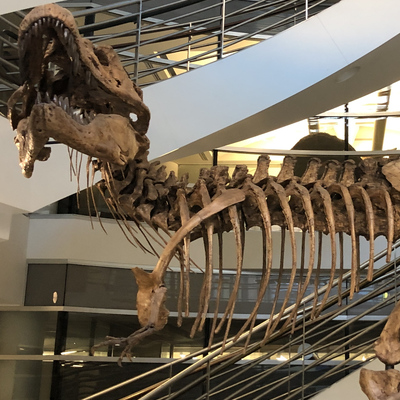}
  \caption{Qualitative evaluation for multi-frame denoising with Gain 20 (best viewed zoomed in). We compare MPFER to 2D-based methods on Spaces (top) and to 3D-based methods on the Real Forward-Facing dataset (bottom).}
  \label{fig:qualitative evaluation}
\end{figure*}

\section{Experiments}

We first consider the Spaces dataset~\cite{flynn2019deepview} and validate our approach on novel view synthesis. We then focus on a denoising setup and perform extensive comparisons to state-of-the-art 2D-based methods. Finally, we compare our approach to the 3D-based multi-frame denoising method of~\cite{pearl2022nan} by replicating their experimental setup on the Real Forward-Facing dataset. In all cases, our method outperforms competitors at a fraction of the computational cost.

\subsection{Spaces}

The Spaces dataset~\cite{flynn2019deepview} consists of 100 indoor and outdoor scenes, captured 5 to 10 times each using a 16-camera rig placed at slightly different locations. 90 scenes are used for training and 10 scenes are held-out for evaluation. The resolution of the images is 480$\times$800.

\paragraph{Novel view synthesis}
We start by replicating the novel view synthesis setup of DeepView~\cite{flynn2019deepview} with four scenarios: one with 12 input views and three with 4 input views. Similarly to DeepView, we use a VGG loss and train our models for 100k steps using the Adam optimizer, with a learning rate of 1.5e-3. We reduce the learning rate to 1.5e-4 after 80k steps, and use a batch size of 4. Memory usage was reported to be a major challenge in DeepView, and all our models are kept at a significantly smaller size to avoid this issue. While DeepView uses a sophisticated strategy to only generate enough of the MPI to render a 32$\times$32 crop in the target image, we use a large patch size of 192 and only apply the loss on the region of the patch that contains more than 80\% of the depths planes after backward warping.  
We compute all metrics by averaging over the validation scenes and target views of each setup, and after cropping a boundary of 16 pixels on all images as done in~\cite{flynn2019deepview}. We compare to DeepView~\cite{flynn2019deepview} and Soft3D~\cite{penner2017soft} by using the synthesised images provided with the Spaces dataset. We also consider three variants of the standard MPI pipeline using the same Unet backbone as our MPFER method and trained in the same conditions, but processing the input PSV in different ways. MPINet implements \emph{Option~1} from \cref{sec:Cross-depth consistency}. The views and depths dimensions of the PSV tensor are stacked along the channel dimension and fed to the Unet backbone to predict the output MPI in one shot. 
MPINet-dw implements \emph{Option~2} from \cref{sec:Cross-depth consistency}. The Unet backbone runs depthwise on slices of the PSV to predict each depth plane of the MPI separately, without communication mechanism across depths. Finally, MPINet-dw-it implements a one-step version of the learned gradient descent algorithm of DeepView. A first estimate of the MPI is predicted by a Unet backbone running depthwise, and this estimate is fed to a second Unet backbone also running depthwise, along with the input PSV and gradient components ({\tt R}), which are PSV-projected current estimates of the target views. 

For our MPFER method and the MPINet ablations, we use a number of depth planes $D=64$ distributed between 100 and 0.5 meters away from the reference camera, placed at the average position of the input views. We use a number of channels $C=8$ and a PSV/MPF upscaling factor $s=1.5$. Since the Unet backbone is not agnostic to the number of input views, we train one version of each model for the setup with 12 input views and one version for the three setups with 4 input views. The results are presented in \Cref{tab:synthesis on spaces}. We observe a clear progression between the performances of MPINet, MPINet-dw and MPINet-dw-it, illustrating the benefit of each design change. Our MPFER method outperforms \mbox{MPINet-dw-it} by up to 4dBs in PSNR at a similar computational complexity, and outperforms DeepView by up to 1.8dB at a fraction of the complexity, clearly motivating the use of a learnt renderer for efficient depth fusion.

\paragraph{Multi-frame denoising}
We now consider a different setup where the inputs are 16 views from one rig position with images degraded with noise, and the targets are the same 16 views denoised. Similarly to previous works~\cite{mildenhall2018burst,xia2020basis,bhat2021deep,pearl2022nan}, we apply synthetic noise with a signal dependent Gaussian distribution $\bm{I}_{vchw} \sim \mathcal{N}\left( \bm{I}_{vchw}^\ast \; , \; \sigma_r^2 + \sigma_s \bm{I}_{vchw}^\ast \right)$
where $\bm{I}$ is the tensor of noisy inputs, $\bm{I}^\ast$ is the ground truth signal, and $\sigma_r$ and $\sigma_s$ are noise parameters that are fixed for each sequence. We focus in particular on challenging scenarios with moderate to high gain levels [4, 8, 16, 20], corresponding to the $\left(\log(\sigma_r), \log(\sigma_s)\right)$ values \mbox{[(-1.44, -1.84)}, (-1.08, -1.48), (-0.72, -1.12), (-0.6, -1.0)] respectively.

We consider two patch-based approaches: VBM4D~\cite{maggioni2012video} and VNLB~\cite{arias2018video}, as well as four state-of-the-art learning-based methods: BPN~\cite{xia2020basis}, BasicVSR~\cite{chan2021basicvsr} and its extension BasicVSR++~\cite{chan2022basicvsr++}, and DeepRep~\cite{bhat2021deep}. To evaluate the influence of the model size and in particular the number of depth planes, we train three MPFER models: \mbox{MPFER-16} with $(D,C,s)\!=\!(16,8,1.)$, MPFER-32 with $(D,C,s)\!=\!(32,16,1.25)$, and MPFER-64 with $(D,C,s)\!=\!(64,8,1.25)$. MPFER-16 has the particularity of using the same number of depth planes as there are input images, meaning that the number of Unet passes per frame to denoise the sequence is \mbox{$(D+V)/V = 2$}. This observation motivates us to perform a comparison with three other architectures, using a strict computational budget of 2 Unet passes per frame. Unet-SF (for Single-Frame) is constituted of two Unet blocks without temporal connection, therefore processing the sequence as a disjoint set of single frames. Unet-BR (for Bidirectional-Recurrent) is constituted of two Unet blocks with bidirectional recurrent connections: the lower Unet processes the sequence in a backward way, and the higher Unet processes the sequence in a feedforward way. Finally, Unet-BR-OF (for Bidirectional-Recurrent with Optical-Flow alignment) is constituted of two Unet blocks with bidirectional recurrent connections, and the recurrent hidden-state is aligned using a SpyNet module, as done in basicVSR~\cite{chan2021basicvsr}. We train all the models in the same conditions as for the novel view synthesis setup, except for the patch size which we increase to 256, and the loss which we replace with a simple L1 loss. During training, we vary the gain level randomly and concatenate an estimate of the standard deviation of the noise to the input, as in~\cite{mildenhall2018burst,xia2020basis,pearl2022nan}. We evaluate on the first rig position of the 10 validation scenes of the Spaces dataset for the 4 gain levels without boundary-cropping, and present the results in \Cref{tab:denoising on spaces}. Each model receives 16 noisy images as input and produces 16 restored images as output, except for BPN and DeepRep which are burst processing methods and only produce one output. For these methods, we choose the view number 6 at the center of the camera rig as the target output, and compare the performances of all methods on this frame. Our MPFER method clearly outperforms all the other methods at a fraction of the computational cost. It performs particularly strongly at high noise levels, with improvements over other methods of more that 2dBs in PSNR. MPFER-16 also performs remarkably well, despite using only 16 depth planes. This suggests that the high representational power of multiplane features allows to significantly reduce the number of depth planes---and therefore the computational cost---compared to standard MPI approaches, which typically use a very high number of planes (80 in the case of DeepView~\cite{flynn2019deepview}, up to 192 in the case of NeX~\cite{wizadwongsa2021nex}). A qualitative evaluation is available in \Cref{fig:qualitative evaluation} (top), and we observe that MPFER is able to reconstruct scenes with much better details. We also present a visualization of multiplane features in \Cref{fig:depth separation}, illustrating how the model learns to separate depths in an unsupervised way.

\subsection{LLFF-N}

The LLFF-N dataset~\cite{pearl2022nan} is a variant of the Real Forward -Facing dataset~\cite{mildenhall2019local} where images are linearized by applying inverse gamma correction and random inverse white balancing, and synthetic noise is applied following the same signal dependent Gaussian distribution as used in the previous section with the six gain levels $[1, 2, 4, 8, 16, 20]$. The dataset contains 35 scenes for training and 8 scenes for testing, and the resolution of the images is 756$\times$1008. 

\paragraph{Denoising}
In this setup, the model receives 8 frames in input: the target frame plus its 7 nearest neighbors in terms of camera distances. We train one MPFER model with $(D,C,s)\!=\!(64,8,1.25)$, using an $\mathcal{L}_1$ loss applied to the target frame. We evaluate on the 43 bursts used in~\cite{pearl2022nan} (every 8th frame of the test set) and present the results in the first half of \Cref{tab:LLFF}. A qualitative evaluation is also available in \Cref{fig:qualitative evaluation} (bottom). To assess the robustness of our method to noisy camera positions, we evaluate it using camera positions computed on clean images (MPFER-C) and on noisy images (MPFER-N) using COLMAP~\cite{schonberger2016structure}. Our method outperforms IBRNet-N and NAN in both scenarios by large margins, but the evaluation using clean camera poses performs significantly better at high noise levels.

\paragraph{Synthesis under noisy conditions}
In this setup, the model receives as input the 8 nearest neighbors to a held-out target. Again, we train one MPFER model with $(D,C,s)\!=\!(64,8,1.25)$, using an $\mathcal{L}_1$ loss applied to the target frame. We evaluate on the same 43 bursts as before and report the results in the second half of \Cref{tab:LLFF}. Our method performs on par with IBRNet and NAN at very low noise levels (close to a pure synthesis problem), and significantly outperforms the other methods at larger noise levels. MPFER only requires $D$ Unet passes to produce an MPF, and 1 Unet pass to render a novel view, which is significantly lighter than IBRNet and NAN. At inference time, the Unet pass requires 0.6 Mflops per pixel, compared to 45 Mflops for IBRNet~\cite{wang2021ibrnet}. A qualitative evaluation is available in \Cref{fig:teaser} for Gain 20.

\begin{figure}[ht]
  \centering
    \makebox[0.326\linewidth]{\centering \footnotesize Noisy input}
    \makebox[0.326\linewidth]{\centering \footnotesize NAN~\cite{pearl2022nan}} \makebox[0.326\linewidth]{\centering \footnotesize MPFER (Ours)}
    
    \includegraphics[width=0.326\linewidth,height=0.326\linewidth]{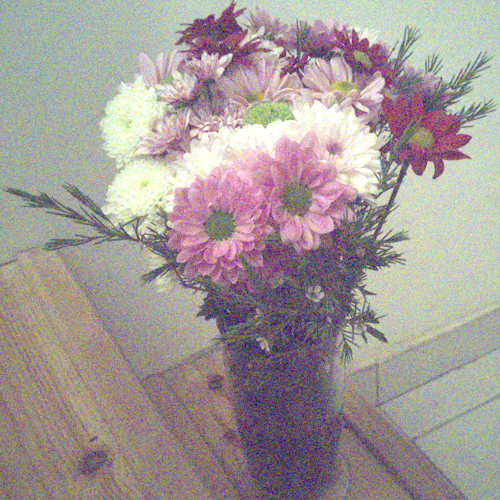} \includegraphics[width=0.326\linewidth,height=0.326\linewidth]{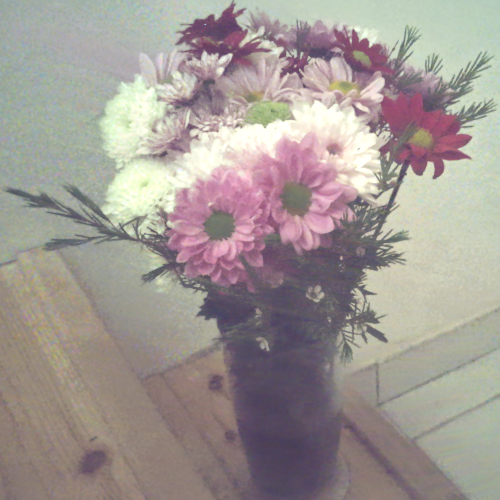} \includegraphics[width=0.326\linewidth,height=0.326\linewidth]{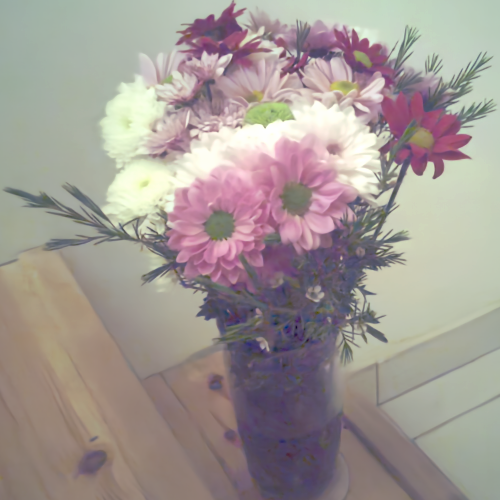}
    \includegraphics[width=0.326\linewidth,height=0.326\linewidth]{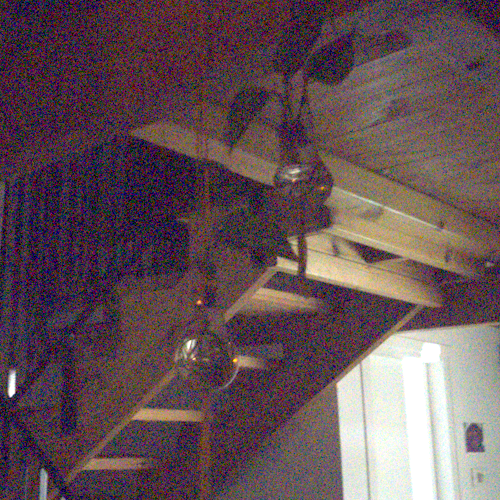} \includegraphics[width=0.326\linewidth,height=0.326\linewidth]{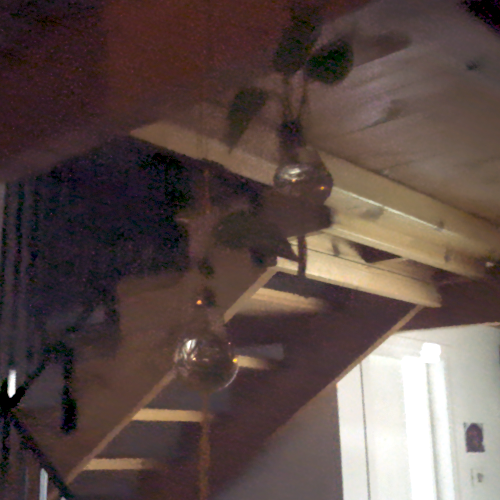} \includegraphics[width=0.326\linewidth,height=0.326\linewidth]{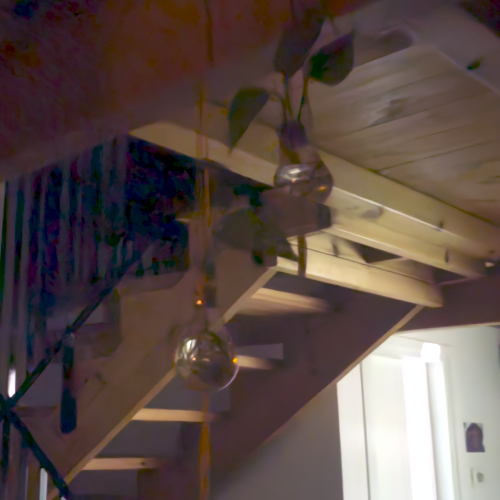}
  \caption{Qualitative evaluation on sequences with real noise from~\cite{pearl2022nan}, captured with a Google Pixel 4.}
  \label{fig:real noise}
\end{figure}

\begin{figure*}[p]
  \centering
   \makebox[0.1111\linewidth]{\scriptsize Noisy input}\makebox[0.1111\linewidth]{\scriptsize Plane 1}\makebox[0.1111\linewidth]{\scriptsize Plane 3}\makebox[0.1111\linewidth]{\scriptsize Plane 5}\makebox[0.1111\linewidth]{\scriptsize Plane 7}\makebox[0.1111\linewidth]{\scriptsize Plane 9}\makebox[0.1111\linewidth]{\scriptsize Plane 11}\makebox[0.1111\linewidth]{\scriptsize Plane 13}\makebox[0.1111\linewidth]{\scriptsize Plane 15}
   \includegraphics[width=0.1111\linewidth]{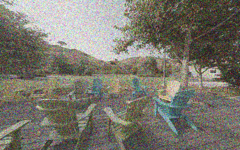}\includegraphics[width=0.1111\linewidth]{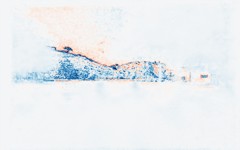}\includegraphics[width=0.1111\linewidth]{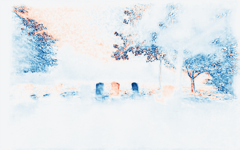}\includegraphics[width=0.1111\linewidth]{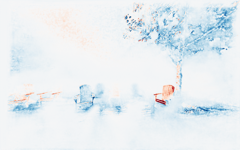}\includegraphics[width=0.1111\linewidth]{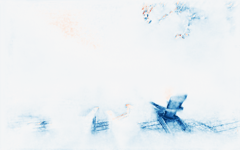}\includegraphics[width=0.1111\linewidth]{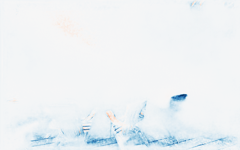}\includegraphics[width=0.1111\linewidth]{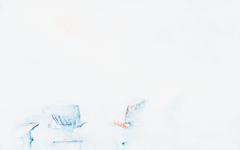}\includegraphics[width=0.1111\linewidth]{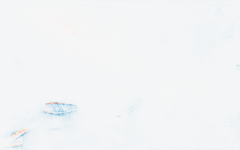}\includegraphics[width=0.1111\linewidth]{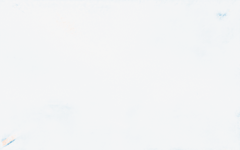}\vspace*{-0.5mm}
   \includegraphics[width=0.1111\linewidth]{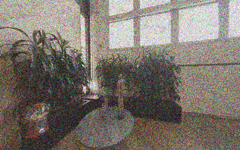}\includegraphics[width=0.1111\linewidth]{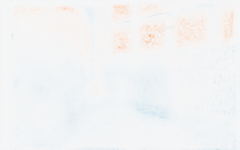}\includegraphics[width=0.1111\linewidth]{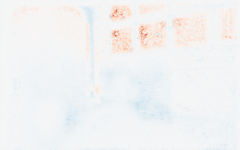}\includegraphics[width=0.1111\linewidth]{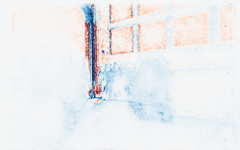}\includegraphics[width=0.1111\linewidth]{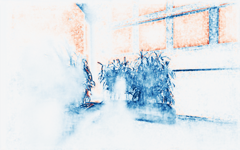}\includegraphics[width=0.1111\linewidth]{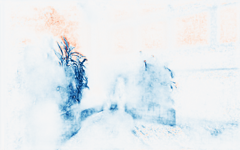}\includegraphics[width=0.1111\linewidth]{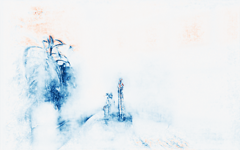}\includegraphics[width=0.1111\linewidth]{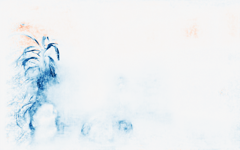}\includegraphics[width=0.1111\linewidth]{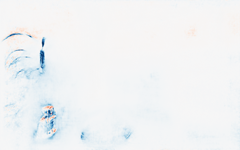}\vspace*{-0.5mm}
   \includegraphics[width=0.1111\linewidth]{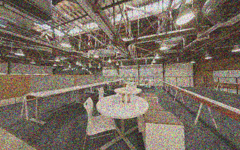}\includegraphics[width=0.1111\linewidth]{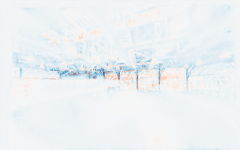}\includegraphics[width=0.1111\linewidth]{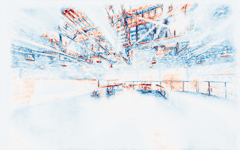}\includegraphics[width=0.1111\linewidth]{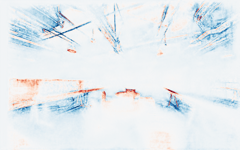}\includegraphics[width=0.1111\linewidth]{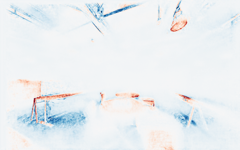}\includegraphics[width=0.1111\linewidth]{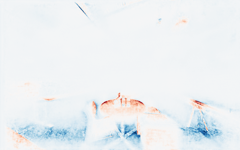}\includegraphics[width=0.1111\linewidth]{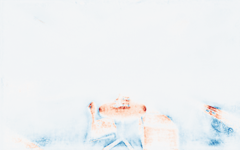}\includegraphics[width=0.1111\linewidth]{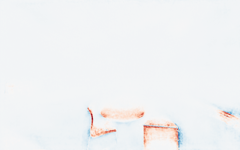}\includegraphics[width=0.1111\linewidth]{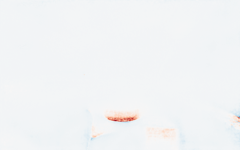}
   \vspace*{-6mm}
   \caption{Visualization of Multiplane Features for 3 scenes. We plot the first 3 channels of planes 1, 3, 5, 7, 9, 11, 13, 15 from MPFER-16. Our method learns to separate depths in an unsupervised way, in this case from a pure denoising problem.}
   \label{fig:depth separation}
\end{figure*}

\paragraph{Low-Light Scenes}
Finally, we qualitatively evaluate our denoising model trained on LLFF-N on sequences with real noise captured with a Google Pixel 4 under low-light conditions. We use the sequences from~\cite{pearl2022nan} and estimate camera poses using COLMAP~\cite{schonberger2016structure} on noisy images. We compare our results to those of~\cite{pearl2022nan}---downloaded from their project page where more baselines can be found---in \Cref{fig:real noise}.

\section{Conclusion}

We proposed to approach multi-frame denoising as a view synthesis problem and argued in favor of using multiplane representations for their low computational cost and generalizability. We introduced a powerful generalization of multiplane images to feature space, and demonstrated its effectiveness in multiple challenging scenarios. 

\setlength{\tabcolsep}{5pt} 
\begin{table*}[p]
\centering
\begin{adjustbox}{width=\textwidth}
\begin{tabular}{@{}lccccccccccccc@{}}
\toprule
\multicolumn{1}{c}{} & \multicolumn{3}{c}{12 input views (dense)} & \multicolumn{3}{c}{4 input views (small)} & \multicolumn{3}{c}{4 input views (medium)} & \multicolumn{3}{c}{4 input views (large)} & \raisebox{-0.1cm}{GFlops@} \\
\cmidrule(lr){2-4}\cmidrule(lr){5-7}\cmidrule(lr){8-10}\cmidrule(lr){11-13}
\multicolumn{1}{c}{} & PSNR$\uparrow$ & SSIM$\uparrow$ & LPIPS$\downarrow$ & PSNR$\uparrow$ & SSIM$\uparrow$ & LPIPS$\downarrow$ & PSNR$\uparrow$ & SSIM$\uparrow$ & LPIPS$\downarrow$ & PSNR$\uparrow$ & SSIM$\uparrow$ & LPIPS$\downarrow$ & \raisebox{0.1cm}{500$\times$800} \\
\hline\hline
Soft3D* & 31.93 & 0.940 & 0.052 & 30.29 & 0.925 & 0.064 & 30.84 & 0.930 & 0.060 & 30.57 & 0.931 & 0.054 & n/a \\
DeepView* & \underline{34.23} & 0.965 & \underline{0.015} & \underline{31.42} & 0.954 & 0.026 & \underline{32.38} & \underline{0.957} & \underline{0.021} & \underline{31.00} & \underline{0.952} & \underline{0.024} & 45800 \\
MPINet & 27.43 & 0.914 & 0.035 & 27.00 & 0.906 & 0.054 & 26.16 & 0.896 & 0.062 & 24.93 & 0.865 & 0.085 & \textbf{450} \\
MPINet-dw & 30.70 & 0.963 & 0.021 & 29.39 & 0.951 & 0.027 & 28.47 & 0.948 & 0.030 & 26.83 & 0.937 & 0.040 & 7890 \\
MPINet-dw-it & 30.85 & \underline{0.966} & 0.017 & 30.22 & \underline{0.955} & \underline{0.024} & 29.37 & 0.953 & 0.026 & 28.00 & 0.943 & 0.034 & 14800 \\
MPFER-64 & \textbf{35.73} & \textbf{0.972} & \textbf{0.012} & \textbf{33.20} & \textbf{0.959} & \textbf{0.018} & \textbf{33.47} & \textbf{0.959} & \textbf{0.018} & \textbf{32.38} & \textbf{0.953} & \textbf{0.021} & 8490 \\
\bottomrule 
\end{tabular}
\end{adjustbox}
\vspace*{-3mm}
\caption{Novel view synthesis on Spaces. All metrics were computed on predicted images with a 16-pixel boundary cropped, as done in~\cite{flynn2019deepview}. Stared methods were evaluated using the predicted images released with the Spaces dataset.}
\label{tab:synthesis on spaces}
\end{table*}

\begin{table*}[p]
\centering
\begin{adjustbox}{width=\textwidth}
\begin{tabular}{@{}lccccccccccccc@{}}
\toprule
\multicolumn{1}{c}{} & \multicolumn{3}{c}{Gain 4} & \multicolumn{3}{c}{Gain 8} & \multicolumn{3}{c}{Gain 16} & \multicolumn{3}{c}{Gain 20} & \raisebox{-0.1cm}{GFlops@} \\
\cmidrule(lr){2-4}\cmidrule(lr){5-7}\cmidrule(lr){8-10}\cmidrule(lr){11-13}
\multicolumn{1}{c}{} & PSNR$\uparrow$ & SSIM$\uparrow$ & LPIPS$\downarrow$ & PSNR$\uparrow$ & SSIM$\uparrow$ & LPIPS$\downarrow$ & PSNR$\uparrow$ & SSIM$\uparrow$ & LPIPS$\downarrow$ & PSNR$\uparrow$ & SSIM$\uparrow$ & LPIPS$\downarrow$ & \raisebox{0.1cm}{500$\times$800} \\
\hline\hline
VBM4D & 32.00 & 0.900 & 0.108 & 29.94 & 0.850 & 0.172 & 27.48 & 0.769 & 0.280 & 26.55 & 0.730 & 0.331 & n/a \\
VNLB & 33.41 & 0.918 & 0.089 & 30.30 & 0.871 & 0.144 & 25.74 & 0.793 & 0.283 & 23.51 & 0.743 & 0.366 & n/a \\
BPN & 34.52 & 0.934 & 0.048 & 32.10 & 0.900 & 0.082 & 29.45 & 0.846 & 0.144 & 28.56 & 0.824 & 0.168 & 810 \\
BasicVSR & 36.87 & 0.959 & 0.027 & 34.52 & 0.937 & 0.049 & 31.73 & 0.898 & 0.095 & 30.68 & 0.879 & 0.119 & 2090 \\ 
BasicVSR++ & 36.98 & 0.959 & 0.026 & 34.66 & 0.938 & 0.045 & 31.97 & 0.902 & 0.083 & 30.92 & 0.883 & 0.102 & 4300 \\
DeepRep & 37.37 & 0.963 & 0.024 & 35.13 & 0.943 & 0.043 & 32.37 & 0.906 & 0.085 & 31.32 & 0.888 & 0.107 & 3230 \\
\midrule
UNet-SF & 35.10 & 0.942 & 0.043 & 32.62 & 0.909 & 0.075 & 29.81 & 0.857 & 0.134 & 28.81 & 0.834 & 0.161 & 440 \\
UNet-BR & 35.19 & 0.943 & 0.040 & 32.67 & 0.912 & 0.070 & 29.90 & 0.861 & 0.124 & 28.94 & 0.840 & 0.148 & 470 \\
UNet-BR-OF & 36.41 & 0.956 & 0.029 & 34.27 & 0.936 & 0.048 & 31.77 & 0.899 & 0.089 & 30.85 & 0.882 & 0.110 & 710 \\
\midrule
MPFER-16 & 37.56 & \underline{0.968} & 0.020 & 35.80 & 0.955 & 0.030 & 33.70 & \underline{0.933} & \underline{0.051} & 32.89 & 0.921 & 0.063 & \textbf{470} \\
MPFER-32 & \underline{37.94} & \textbf{0.970} & \underline{0.019} & \underline{36.17} & \underline{0.958} & \underline{0.028} & \underline{33.99} & \textbf{0.936} & \textbf{0.047} & \underline{33.14} & \underline{0.924} & \underline{0.058} & 1210 \\
MPFER-64 & \textbf{38.00} & \textbf{0.970} & \textbf{0.018} & \textbf{36.25} & \textbf{0.959} & \textbf{0.027} & \textbf{34.08} & \textbf{0.936} & \textbf{0.047} & \textbf{33.23} & \textbf{0.925} & \textbf{0.057} & 1810 \\
\bottomrule
\end{tabular}
\end{adjustbox}
\vspace*{-3mm}
\caption{Denoising on Spaces. All metrics were computed on frame 6, as BPN and DeepRep are burst-denoising methods producing only one output. For the other methods, average performances over the entire sequence are provided in Supplementary Material.}
\label{tab:denoising on spaces}
\end{table*}

\setlength{\tabcolsep}{1pt}
\begin{table*}[p]
\centering
\begin{adjustbox}{width=\textwidth}
\begin{tabular}{@{}lcccccccccccccccccc@{}}
\toprule
\multicolumn{1}{c}{} & \multicolumn{3}{c}{Gain 1} & \multicolumn{3}{c}{Gain 2} & \multicolumn{3}{c}{Gain 4} & \multicolumn{3}{c}{Gain 8} & \multicolumn{3}{c}{Gain 16} & \multicolumn{3}{c}{Gain 20} \\
\cmidrule(lr){2-4}\cmidrule(lr){5-7}\cmidrule(lr){8-10}\cmidrule(lr){11-13}\cmidrule(lr){14-16}\cmidrule(lr){17-19}
\multicolumn{1}{c}{} & PSNR$\uparrow$ & SSIM$\uparrow$ & LPIPS$\downarrow$ & PSNR$\uparrow$ & SSIM$\uparrow$ & LPIPS$\downarrow$ & PSNR$\uparrow$ & SSIM$\uparrow$ & LPIPS$\downarrow$ & PSNR$\uparrow$ & SSIM$\uparrow$ & LPIPS$\downarrow$ & PSNR$\uparrow$ & SSIM$\uparrow$ & LPIPS$\downarrow$ & PSNR$\uparrow$ & SSIM$\uparrow$ & LPIPS$\downarrow$ \\
\midrule
 & \multicolumn{18}{c}{\emph{Denoising of Synthetic Noise}} \\
IBRNet-N* & 33.50 & 0.915 & 0.039 & 31.29 & 0.877 & 0.070 & 29.01 & 0.822 & 0.123 & 26.57 & 0.741 & 0.210 & 24.19 & 0.634 & 0.331 & 23.40 & 0.591 & 0.380 \\
NAN* & 35.84 & 0.955 & 0.018 & 33.67 & 0.930 & 0.034 & 31.26 & 0.892 & 0.068 & 28.64 & 0.834 & 0.132 & 25.95 & 0.749 & 0.231 & 25.07 & 0.715 & 0.271 \\
MPFER-N & \underline{37.90} & \underline{0.969} & \underline{0.013} & \underline{35.61} & \underline{0.951} & \underline{0.025} & \underline{33.02} & \underline{0.921} & \underline{0.048} & \underline{30.21} & \underline{0.872} & \underline{0.091} & \underline{27.24} & \underline{0.797} & \underline{0.164} & \underline{26.23} & \underline{0.765} & \underline{0.198} \\
MPFER-C & \textbf{38.06} & \textbf{0.971} & \textbf{0.011} & \textbf{35.95} & \textbf{0.956} & \textbf{0.020} & \textbf{33.65} & \textbf{0.934} & \textbf{0.036} & \textbf{31.21} & \textbf{0.898} & \textbf{0.065} & \textbf{28.61} & \textbf{0.843} & \textbf{0.115} & \textbf{27.71} & \textbf{0.819} & \textbf{0.138} \\
\midrule
 & \multicolumn{18}{c}{\emph{Novel View Synthesis Under Noisy Conditions}} \\
IBRNet* & \textbf{24.53} & 0.774 & \underline{0.135} & 24.20 & 0.730 & 0.159 & 23.44 & 0.653 & 0.217 & 22.02 & 0.536 & 0.327 & 19.76 & 0.377 & 0.492 & 18.80 & 0.319 & 0.553 \\
IBRNet-N* & 23.86 & 0.763 & 0.170 & 23.73 & 0.744 & 0.178 & 23.38 & 0.703 & 0.208 & 22.68 & 0.638 & 0.275 & 21.67 & 0.549 & 0.377 & 21.29 & 0.514 & 0.418 \\
NAN* & \underline{24.52} & \textbf{0.799} & \textbf{0.132} & \underline{24.41} & \underline{0.787} & \textbf{0.145} & \underline{24.18} & \underline{0.765} & \underline{0.171} & \underline{23.70} & \underline{0.726} & \underline{0.221} & \underline{22.79} & \underline{0.666} & \underline{0.305} & \underline{22.37} & \underline{0.641} & \underline{0.342} \\
MPFER & \underline{24.52} & \underline{0.798} & 0.157 & \textbf{24.51} & \textbf{0.796} & \underline{0.158} & \textbf{24.47} & \textbf{0.789} & \textbf{0.164} & \textbf{24.33} & \textbf{0.775} & \textbf{0.178} & \textbf{23.94} & \textbf{0.746} & \textbf{0.212} & \textbf{23.72} & \textbf{0.731} & \textbf{0.230}  \\
\bottomrule
\end{tabular}
\end{adjustbox}
\vspace*{-3mm}
\caption{LLFF-N. We consider the two scenarios introduced in~\cite{pearl2022nan}: Denoising of Synthetic Noise, where the noisy target is accessible, and Novel View Synthesis Under Noisy Conditions, where the noisy target is held-out. The numbers for the stared methods correspond to Figure~9 in~\cite{pearl2022nan}, and were provided by the authors.}
\label{tab:LLFF}
\end{table*}

\cleardoublepage
{\small
\bibliographystyle{ieee_fullname}
\bibliography{references}
}

\clearpage
\appendix

\section{Network architecture}

\begin{figure}[ht]
  \centering
  \includegraphics[width=0.9\linewidth]{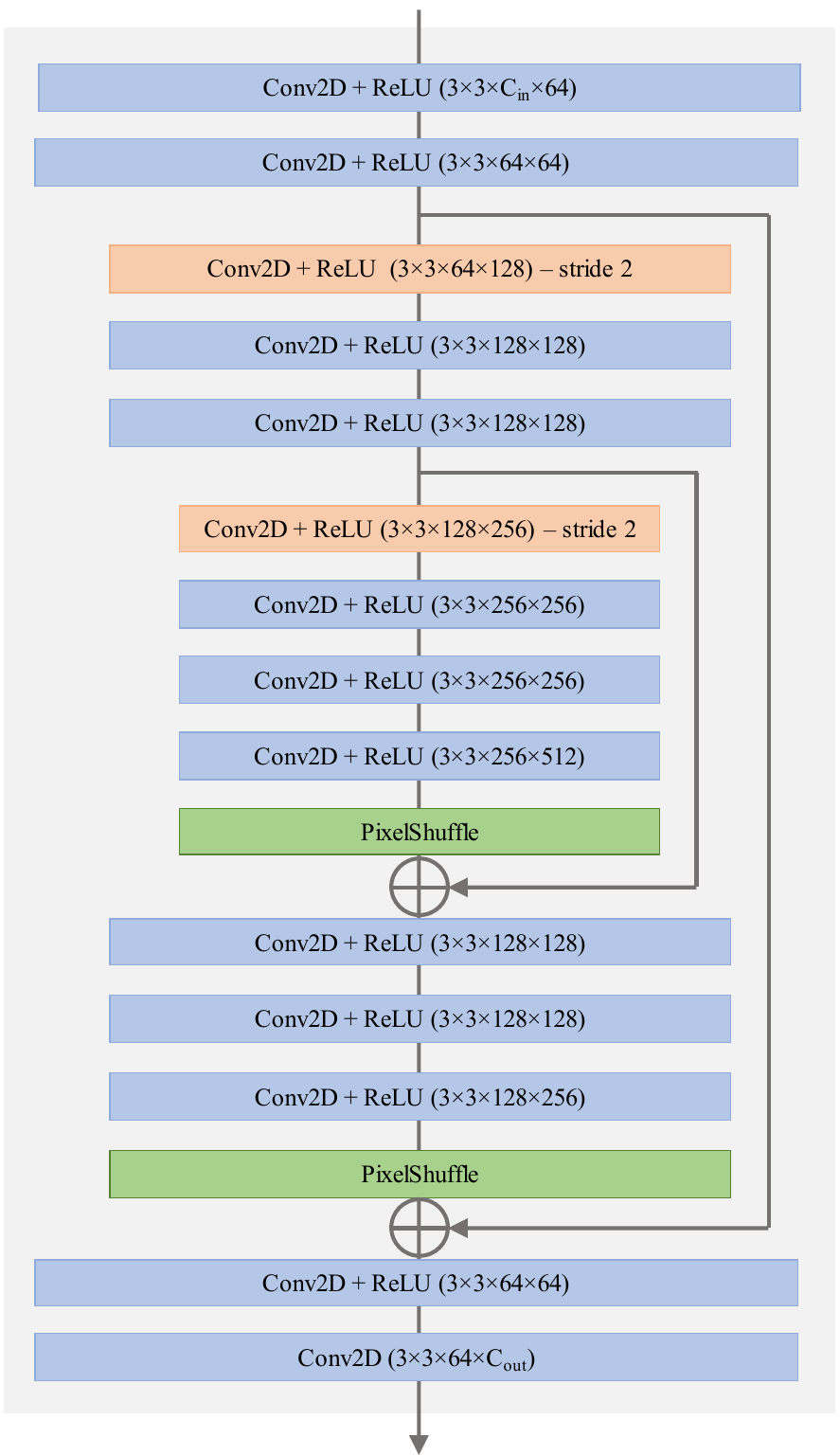}
  \caption{Unet architecture used for the Encoder and for the Renderer in all our MPFER experiments.}
  \label{fig:unet}
\end{figure}

The architecture of our Multiplane Feature Encoder-Renderer (MPFER) is described in the main paper and illustrated in Figure~3. The Encoder and the Renderer consist in two identical Unets with a base of 64 channels, illustrated in more details in \Cref{fig:unet}.

\section{Average metrics}

\begin{figure}[ht]
    \makebox[0.19\linewidth]{\footnotesize Noisy}\hfill
    \includegraphics[width=0.75\linewidth]{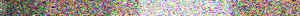}
    
    \makebox[0.19\linewidth]{\footnotesize BPN}\hfill
    \includegraphics[width=0.75\linewidth]{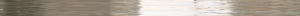}
    
    \makebox[0.19\linewidth]{\footnotesize BasicVSR++}\hfill
    \includegraphics[width=0.75\linewidth]{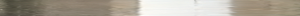}
    
    \makebox[0.19\linewidth]{\footnotesize DeepRep}\hfill
    \includegraphics[width=0.75\linewidth]{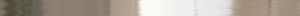}
    
    \makebox[0.23\linewidth]{\footnotesize MPFER-64 (ours)}\hfill 
    \includegraphics[width=0.75\linewidth]{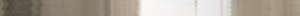}
    
    \makebox[0.19\linewidth]{\footnotesize Ground truth}\hfill
    \includegraphics[width=0.75\linewidth]{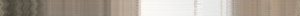}
  \caption{V$\times$W slices computed on scene 52 of Spaces.}
  \label{fig:slices}
\end{figure}

In Section~4.1, Table~2, We compare our MPFER model to various 2D-based video restoration methods for denoising of synthetic noise on the Spaces dataset. Two of the methods we consider, BPN~\cite{xia2020basis} and DeepRep~\cite{bhat2021deep}, are burst denoising methods producing only one denoised output for the entire set of noisy inputs. By default, we chose this output to be frame number~6 at the center of the camera rig and compared the performances of all the methods on that frame. However, MPFER as well as BasicVSR~\cite{chan2021basicvsr} and BasicVSR++~\cite{chan2022basicvsr++}, are multi-frame denoising methods producing one denoised output per noisy input. We compare their average performances over the 16 frames of the validation sequences in \Cref{tab:average metrics}. We see that the overall performances are comparable to those on frame~6 from Table~2. In particular, MPFER outperforms all other methods by large margins on all noise levels. To qualitatively evaluate the cross-view consistency of different methods, we also plot V$\times$W slices computed on scene 52 in \Cref{fig:slices}. We run BPN and DeepRep 16 times (once per frame) to obtain these profiles. Our method qualitatively matches the ground-truth better than other methods.

\section{Ablations}

Our MPFER method depends on three hyperparameters: the number of depth planes $D$, the number of channels in the multiplane representation $C$, and the upscaling factor of the PSV/MPF representation $s$. In Table~2 of the main paper, we evaluated the influence of model size by varying these three hyperparameters simultaneously. We now evaluate the influence of each hyperparameter independently in \Cref{tab:ablations}. We see that the performance of the method increases with $D$, $C$ and $s$, and so does the computational complexity. Interestingly, the performance improvement is higher when $C$ increases from 4 to 16 (+0.4dB at Gain 20), than when $D$ increases from 16 to 64 (+0.23dB at Gain 20), while the increase in computational complexity is significantly lower ($\times 1.33$ vs $\times 2.97$ respectively). This observation confirms that multiplane features are inherently more powerful representations than multiplane images, allowing to perform efficient 3D-based video restoration with fewer depth planes.

\section{Qualitative evaluations}

We consider 4 experimental setups in the main paper: (1) Novel View Synthesis on Spaces, (2) Denoising on Spaces, (3) Denoising on the Real Forward Facing dataset, and (4) Novel View Synthesis under Noise Conditions on the Real Forward Facing dataset. We present some visual comparisons with state-of-the-art methods for the setups (2) and (3) in Figure~4, and for the setup (4) in Figure~1. We present some additional visual comparisons for the setup (1) in \Cref{fig:qualitative evaluation} here.

\setlength{\tabcolsep}{5pt} 
\begin{table*}[ht]
\centering
\begin{adjustbox}{width=\textwidth}
\begin{tabular}{@{}lccccccccccccc@{}}
\toprule
\multicolumn{1}{c}{} & \multicolumn{3}{c}{Gain 4} & \multicolumn{3}{c}{Gain 8} & \multicolumn{3}{c}{Gain 16} & \multicolumn{3}{c}{Gain 20} & \raisebox{-0.1cm}{GFlops@} \\
\cmidrule(lr){2-4}\cmidrule(lr){5-7}\cmidrule(lr){8-10}\cmidrule(lr){11-13}
\multicolumn{1}{c}{} & PSNR$\uparrow$ & SSIM$\uparrow$ & LPIPS$\downarrow$ & PSNR$\uparrow$ & SSIM$\uparrow$ & LPIPS$\downarrow$ & PSNR$\uparrow$ & SSIM$\uparrow$ & LPIPS$\downarrow$ & PSNR$\uparrow$ & SSIM$\uparrow$ & LPIPS$\downarrow$ & \raisebox{0.1cm}{500$\times$800} \\
\hline\hline
VBM4D & 32.30 & 0.90 & n/a & 30.12 & 0.849 & n/a & 27.53 & 0.763 & n/a & 26.58 & 0.723 & n/a & n/a \\
VNLB & 33.41 & 0.917 & 0.089 & 30.31 & 0.869 & 0.144 & 25.79 & 0.794 & 0.279 & 23.58 & 0.746 & 0.363 & n/a \\
BasicVSR & 36.86 & 0.957 & 0.029 & 34.45 & 0.935 & 0.052 & 31.62 & 0.895 & 0.099 & 30.59 & 0.875 & 0.124 & 2090 \\ 
BasicVSR++ & 36.81 & 0.957 & 0.030 & 34.39 & 0.934 & 0.051 & 31.62 & 0.895 & 0.091 & 30.60 & 0.875 & 0.111 & 4300 \\
\midrule
UNet-SF & 35.15 & 0.942 & 0.042 & 32.67 & 0.910 & 0.074 & 29.86 & 0.857 & 0.134 & 28.87 & 0.834 & 0.160 & \textbf{440} \\
UNet-BR & 35.23 & 0.943 & 0.040 & 32.72 & 0.912 & 0.070 & 29.97 & 0.861 & 0.124 & 29.02 & 0.840 & 0.148 & \textbf{470} \\
UNet-BR-OF & 36.37 & 0.955 & 0.029 & 34.18 & 0.934 & 0.049 & 31.65 & 0.896 & 0.091 & 30.71 & 0.878 & 0.112 & 710 \\
\midrule
MPFER-16 & 37.20 & 0.965 & \underline{0.021} & 35.37 & 0.952 & \underline{0.033} & 33.22 & 0.927 & 0.055 & 32.41 & 0.915 & 0.067 & \textbf{470} \\
MPFER-32 & \underline{37.52} & \underline{0.967} & \textbf{0.020} & \underline{35.69} & \underline{0.954} & \textbf{0.030} & \underline{33.50} & \underline{0.931} & \underline{0.051} & \underline{32.66} & \underline{0.919} & \underline{0.063} & 1210 \\
MPFER-64 & \textbf{37.60} & \textbf{0.968} & \textbf{0.020} & \textbf{35.78} & \textbf{0.955} & \textbf{0.030} & \textbf{33.58} & \textbf{0.932} & \textbf{0.050} & \textbf{32.74} & \textbf{0.920} & \textbf{0.061} & 1810 \\
\bottomrule
\end{tabular}
\end{adjustbox}
\vspace*{-2mm}
\caption{Denoising on Spaces. Average metrics over the 16 frames in the validation sequences. Best results in \textbf{bold}, second best \underline{underlined}.}
\label{tab:average metrics}
\end{table*}

\setlength{\tabcolsep}{5pt} 
\begin{table*}[ht]
\centering
\begin{adjustbox}{width=\textwidth}
\begin{tabular}{@{}lccccccccccccc@{}}
\toprule
\multicolumn{1}{c}{} & \multicolumn{3}{c}{Gain 4} & \multicolumn{3}{c}{Gain 8} & \multicolumn{3}{c}{Gain 16} & \multicolumn{3}{c}{Gain 20} & \raisebox{-0.1cm}{GFlops@} \\
\cmidrule(lr){2-4}\cmidrule(lr){5-7}\cmidrule(lr){8-10}\cmidrule(lr){11-13}
\multicolumn{1}{l}{$(D,C,s)$} & PSNR$\uparrow$ & SSIM$\uparrow$ & LPIPS$\downarrow$ & PSNR$\uparrow$ & SSIM$\uparrow$ & LPIPS$\downarrow$ & PSNR$\uparrow$ & SSIM$\uparrow$ & LPIPS$\downarrow$ & PSNR$\uparrow$ & SSIM$\uparrow$ & LPIPS$\downarrow$ & \raisebox{0.1cm}{500$\times$800} \\
\hline\hline
\multicolumn{14}{c}{\emph{Influence of the number of depth planes}} \\
$(16,8,1.25)$ & 37.75 & 0.969 & 0.019 & 35.98 & 0.957 & 0.028 & 33.83 & 0.934 & 0.049 & 33.00 & 0.922 & 0.061 & 610 \\
$(32,8,1.25)$ & 37.86 & 0.970 & 0.018 & 36.11 & 0.958 & 0.027 & 33.95 & 0.935 & 0.047 & 33.10 & 0.924 & 0.059 & 1010 \\
$(64,8,1.25)$ & 38.00 & 0.970 & 0.018 & 36.25 & 0.959 & 0.027 & 34.08 & 0.936 & 0.047 & 33.23 & 0.925 & 0.057 & 1810 \\
\midrule
\multicolumn{14}{c}{\emph{Influence of the number of channels in the multiplane representation}} \\
$(32,4,1.25)$ & 37.64 & 0.969 & 0.021 & 35.81 & 0.957 & 0.031 & 33.59 & 0.935 & 0.051 & 32.74 & 0.923 & 0.062 & 910 \\
$(32,8,1.25)$ & 37.86 & 0.970 & 0.018 & 36.11 & 0.958 & 0.027 & 33.95 & 0.935 & 0.047 & 33.10 & 0.924 & 0.059 & 1010 \\
$(32,16,1.25)$ & 37.94 & 0.970 & 0.019 & 36.17 & 0.958 & 0.028 & 33.99 & 0.936 & 0.047 & 33.14 & 0.924 & 0.058 & 1210\\
\midrule
\multicolumn{14}{c}{\emph{Influence of the upscaling factor}} \\
$(16,8,1.0)$ & 37.56 & 0.968 & 0.020 & 35.80 & 0.955 & 0.030 & 33.70 & 0.933 & 0.051 & 32.89 & 0.921 & 0.063 & 470\\
$(16,8,1.25)$ & 37.75 & 0.969 & 0.019 & 35.98 & 0.957 & 0.028 & 33.83 & 0.934 & 0.049 & 33.00 & 0.922 & 0.061 & 610 \\
$(16,8,1.5)$ & 37.86 & 0.969 & 0.019 & 36.08 & 0.957 & 0.029 & 33.92 & 0.935 & 0.050 & 33.08 & 0.923 & 0.061 & 780 \\
\bottomrule
\end{tabular}
\end{adjustbox}
\vspace*{-2mm}
\caption{Denoising on Spaces. Influence of hyperparameters $(D,C,s)$: number of depth planes, number of channels in the multiplane representation, upscaling factor.}
\label{tab:ablations}
\end{table*}

\begin{figure*}[ht]
  \centering
    \makebox[0.326\linewidth]{MPINet}\hfill
    \makebox[0.326\linewidth]{MPINet-dw}\hfill
    \makebox[0.326\linewidth]{MPINet-dw-it}
    \includegraphics[width=0.163\linewidth]{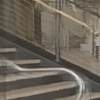}\includegraphics[width=0.163\linewidth]{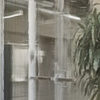}\hfill \includegraphics[width=0.163\linewidth]{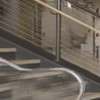}\includegraphics[width=0.163\linewidth]{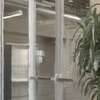}\hfill \includegraphics[width=0.163\linewidth]{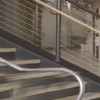}\includegraphics[width=0.163\linewidth]{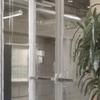}

    \makebox[0.326\linewidth]{DeepView}\hfill
    \makebox[0.326\linewidth]{MPFER-64 (ours)}\hfill
    \makebox[0.326\linewidth]{Ground truth}
    \includegraphics[width=0.163\linewidth]{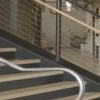}\includegraphics[width=0.163\linewidth]{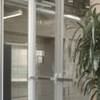}\hfill \includegraphics[width=0.163\linewidth]{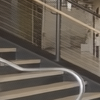}\includegraphics[width=0.163\linewidth]{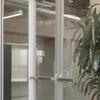}\hfill \includegraphics[width=0.163\linewidth]{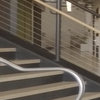}\includegraphics[width=0.163\linewidth]{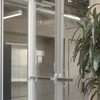}

  \caption{Qualitative evaluation for novel view synthesis on Spaces (best viewed zoomed in).}
  \label{fig:qualitative evaluation}
\end{figure*}

\end{document}